\documentclass[11pt,a4paper]{article}

\usepackage[utf8]{inputenc} 
\usepackage[T1]{fontenc}    

\usepackage[hyperref]{acl2019}
\usepackage{times}
\usepackage{latexsym}
\usepackage{microtype}
\usepackage{url}

\aclfinalcopy 
\def\aclpaperid{141} 

\usepackage[capitalise]{cleveref}
\usepackage{booktabs}
\usepackage{graphicx}
\usepackage{multirow}
\usepackage{subcaption}
\usepackage{amsmath}
\usepackage{float}
\usepackage[super]{nth}
\usepackage{xspace}

\usepackage{xcolor}
\usepackage[normalem]{ulem}

\def\VRdel#1{\bgroup\markoverwith{\textcolor{magenta}{\rule[0.5ex]{2pt}{1pt}}}\ULon{#1}}

\definecolor{darkgreen}{rgb}{0.0, 0.5, 0.0}

\def\ODdel#1{\bgroup\markoverwith{\textcolor{darkgreen}{\rule[0.5ex]{2pt}{1pt}}}\ULon{#1}}

\def\onedim{\emph{one-dim}\xspace}
\def\multidim{\emph{multi-dim}\xspace}
\def\transfixed{\emph{trans-fixed}\xspace}
\def\transadapt{\emph{trans-adapt}\xspace}

\newcommand{\ignore}[1]{}

\title{User Evaluation of a Multi-dimensional Statistical Dialogue System}

\author{Simon Keizer,$^{\ast}$ Ondřej Dušek,$^{\ast\dag}$ Xingkun Liu$^{\ast}$ \and Verena Rieser$^{\ast}$ \\
  $^{\ast}$Interaction Lab, Heriot-Watt University, Edinburgh, Scotland, UK \\
  $^{\dag}$Charles University, Faculty of Mathematics and Physics, Prague, Czechia \\
  \texttt{keizer.simon@gmail.com, odusek@ufal.mff.cuni.cz,}\\ \texttt{\{x.liu,v.t.rieser\}@hw.ac.uk}}

\date{}

\begin{document}

\maketitle

\begin{abstract}
We present the first complete spoken dialogue system driven by a \emph{multi-dimensional} statistical dialogue manager.  This framework has been shown to substantially reduce data needs by leveraging domain-independent dimensions, such as social obligations or feedback, which (as we show) can be  transferred between domains.
In this paper, we conduct a user study and show that the performance of a multi-dimensional system, which can be adapted from a source domain, is equivalent to that of a one-dimensional baseline, which can only be trained from scratch.
\end{abstract}

\section{Introduction}\label{sec:intro}

Data-driven approaches to spoken dialogue systems (SDS) are limited by their reliance on substantial amounts of annotated data in the target domain.  This can be addressed by considering transfer learning techniques, e.g.\  \cite{Taylor:2009ur}, in which data from a source domain is leveraged to improve learning in a target domain.  In particular, domain adaptation has been used in the context of dialogue systems \cite{Gasic-etal-CSL2016, Wang:2015tp, wen-etal-2016-multi}, focusing on identifying and exploiting similarities between domain ontologies in slot-filling tasks.  

In contrast to this previous work, we take a \emph{multi-dimensional} approach, which combines machine learning with linguistic theory.  Following \citet{Bunt:2011et}, we exploit the linguistic phenomenon that utterances serve more than one function in a conversation, i.e.\ they have more than one \emph{dimension} (see Section~\ref{sec:mdm}).\footnote{See also \url{https://dit.uvt.nl/}.}  For example, the utterance ``On what date would you like to fly to London?''\ both asks a task-oriented question, and provides feedback about understanding the requested destination.  We take advantage of this phenomenon by training separate, fully-statistical dialogue models for each dimension and generating system responses along multiple dimensions simultaneously.
Such an SDS thus has the potential to adapt more efficiently to new domains by exploiting previously trained policies of the domain-independent dimensions, such as feedback and social conventions.

Previous implementations of multi-di\-men\-sio\-nal SDSs were mostly handcrafted \cite{Schooten_ea-2005,Petukhova_ea-lrec-2016}.
\citet{Keizer_Rieser:2017} were the first to present a statistical multi-dimensional dialogue manager (DM).  Their results suggest an up to 80\% reduction in data: a task success rate of over 90\% can be achieved after only 2,000 dialogues when using pre-trained policies, whereas at least 10,000 dialogues are required without pre-training.  In comparison, \citet{Gasic-etal-CSL2016} achieve similar success rates for in-domain systems trained on 5,000 dialogues. However, \citeauthor{Keizer_Rieser:2017}'s findings are only tested in simulation.

In this paper, we present the first complete statistical SDS with multi-dimensional DM, and the first crowdsourced human user evaluation of this type of system, comparing a one-dimensional baseline and three multi-dimensional variants, using a novel web-based setup.  A novel aspect of our statistical analysis is testing for \emph{equivalence}.  The four system variants were designed in such a way that we would expect their performance levels to be indistinguishable when using fully trained policies.  Should the data provide statistical evidence for this, the multi-dimensional variants can be preferred due to their inherent potential for domain transfer.

\section{A Multi-dimensional Dialogue Manager}\label{sec:mdm}

Our DM is a partially-observable Markov decision process \cite[POMDP;][]{young_pomdp-based_2013} which takes as input an n-best list of dialogue act hypotheses, updates the dialogue state and then selects a response in the form of one or more dialogue acts.
Rather than selecting a single action from one set of possible actions, our DM consists of multiple \emph{dialogue act agents}, each of which selects an action from a separate action set, associated with one dimension.  These action sets are based on three of the ten dimensions defined in the ISO standard for dialogue act annotation \cite{ISO-SemAnnot}: Task (e.g. recommending a restaurant), AutoFeedback (e.g. asking the user to repeat/rephrase after a processing problem), and Social Obligations Management (SOM; e.g.\ responding to the user saying goodbye).  
These dimensions were considered to be the most important for supporting the kind of task-oriented dialogues targeted (see \cref{fig:multi-dim} for an example). While the Task dimension is domain-specific, AutoFeedback and SOM are applicable across domains.

\begin{figure}[tb]
\begin{center}
\footnotesize
\begin{tabular}{@{}l@{\;}l@{}}
\toprule
\textbf{Usr}: & \textsl{\textbf{Hi, I need a \underline{Thai} restaurant in the \underline{city centre}}} \\[1mm]
\multicolumn{2}{l}{\hspace{3mm}\textsc{Social: Greet; Task: Inform; Turn: Release}} \\[2mm]
\midrule
\textbf{Sys}: & \textsl{\textbf{Okay, let me see, \dots}} \\
\multicolumn{2}{l}{\hspace{3mm}\textsc{Turn: Take; AutoFeedback: autoPositive}} \\
\multicolumn{2}{l}{\hspace{3mm}\textsc{Time: Pausing; Task: InformSearch}} \\[2mm]
\textbf{Sys}: & \textsl{\textbf{Bangkok City is a \underline{Thai} restaurant};} \\[1mm]
              & \hspace{3mm}\textsl{\textbf{it is in the \underline{city centre}}} \\[2mm]
\multicolumn{2}{l}{\hspace{3mm}\textsc{AutoFeedback: Inform; Task: Inform}} \\
\bottomrule
\end{tabular}
\end{center}
\caption{An example of multiple dimensions in a dialogue: the user both greets the system and asks for a cheap Indian restaurant, before releasing the turn; the system then takes the turn while giving positive feedback, and indicates that it needs some time to retrieve the requested information; in the second part the system both provides this information and gives feedback about understanding the user's question (underlined).}
\label{fig:multi-dim}
\end{figure}

Training the statistical DM on these three dimensions involves optimising three policies in parallel.  A set of priority rules is used to combine the output of these policies into a single system response.
The key advantage of such a design is that the domain-independent policies (AutoFeedback and SOM) can be transferred and adapted to a new domain, leaving only the Task policy to be trained from scratch.
In our previous work \cite{Keizer_Rieser:2017}, we have shown that a multi-dimensional DM with pre-trained policies reaches higher performance levels during the early stages of training.
Here, we take an important step in confirming this advantage in a real user study.

Our framework currently supports information-seeking domains, such as recommending restaurants or hotels based on the user's preferences.  The domains are specified in terms of an ontology (describing slots such as price range and cuisine) and a database.  
Our domains are presented in \cref{tab:doms}.  We use restaurant information as target domain, but two of the system variants were trained for the hotels domain (source) and then adapted to the restaurant domain.
\begin{table}[t]
\centering \small
\begin{tabular}{lcc}
\toprule
& {\bf Restaurants} & {\bf Hotels} \\
\midrule
\#venues      & 149 & 39 \\
\#slots       & 4 & 5 \\
shared slots  & \multicolumn{2}{c}{\tt pricerange, area, near} \\
other slots   & {\tt cuisine} & {\tt type, rating} \\
\bottomrule
\end{tabular}
\caption{Overview of task domains.}\label{tab:doms}
\vspace{-2mm}
\end{table}

\subsection{Model Variants}\label{sec:models}

For the evaluation, we follow \citet{Keizer_Rieser:2017}'s four DM variants and training regime:
The one-dimensional \onedim baseline system contains a single dialogue act agent (\textsc{All}) and the corresponding policy was trained from scratch in the target domain.
The multi-dimensional systems use three dialogue act agents, one of which is domain-specific (\textsc{Task}) and the other two domain-general (\textsc{AutoFeedback} and \textsc{SOM}).  For the base \multidim system, the three policies are trained from scratch in the target domain, whereas the \transfixed and \transadapt variants employ transfer learning \cite{Pan:2010dm,torrey_transfer_2010}: only the task-specific policy is trained from scratch and the two domain-general policies are previously trained in the source domain. 
For \transfixed, the pre-trained policies are kept fixed during training in the target domain, whilst for \transadapt, these are further trained in the target domain. 
The four fully trained DM versions are outlined in \cref{tab:systems}.

\subsection{Training Details}\label{sec:training}

All policies are optimised in simulation using multi-agent reinforcement learning with linear value function approximation, based on a single reward signal shared between the agents.
\footnote{%
The reward function, shared among the agents/\hspace{0mm}dimensions, was the following: (i) a reward of +80 upon task completion, (ii) a penalty of -1 for each turn, (iii) a reward of +3 when responding appropriately to a social act, and (iv) a penalty of -5 when not signalling a perception or interpretation level processing problem to the user when it occurred.

For each of the four DM versions, 5 training runs over 60k dialogues were carried out, resulting in a pool of 5 fully trained policies.
}
To train all systems, we use the agenda-based user simulator of \citet{Keizer_Rieser:2017}, which is based on \cite{Schatzmann_ea-SLT-2007}, along with the following error model:
In addition to creating an n-best list of user dialogue act hypotheses from the `true' user act, we also occasionally insert so-called `processing problems', at the levels of perception (no ASR results received) or interpretation (ASR successful, but no NLU results received).  We simulate a perception problem with 10\% probability, and in case of no perception problem (90\%), we simulate an interpretation problem with 10\% probability; only in case no processing problems are generated (81\%), an n-best list of dialogue act hypotheses is generated.  Following \citet{Thomson_ea-SLT-2012}, the n-best lists are populated by taking the true user act and distorting it at a given semantic error rate for each of the positions, after which semantically equivalent hypotheses are merged.  Based on the error rate, a Dirichlet distribution is used to generate confidence scores for the n-best list (resulting in a semantic top accuracy equal to the error rate), interpreted as probabilities by the DM when updating its user goal belief state.\footnote{The n-best size was set to 3 and the error rate was set to 30\% for the target domain (restaurants) and 20\% for the source domain (hotels).}

In order to correctly interpret the evaluation results, note that in the current setup, the \onedim system serves as an upper bound baseline system, as it needs no coordination between different agents during training whilst generating (by construction) the same range of actions as the multi-dimensional systems.  This is ensured by a set of priority heuristics which map action combinations to single acts.\footnote{E.g.\ if the Task agent generates a recommendation action and the AutoFeedback agent generates a negative feedback action, the latter gets priority and the former is cancelled.}

\begin{table*}[htb]
\small
\centering
\begin{tabular}{lllll}
\toprule
{\bf Dialogue Act Agent} & {\bf \onedim} & {\bf \multidim} & {\bf \transfixed} & {\bf \transadapt} \\
\midrule
\textsc{All}  & {\it source}: --      & --              & --              & -- \\
              & {\it target}: trained & --              & --              & -- \\
\midrule
\textsc{Task} & --       & {\it source}: --          & {\it source}: --             & {\it source}: -- \\
              & --       & {\it target}: trained     & {\it target}: trained        & {\it target}: trained \\[1mm]
\textsc{AutoFeedback} & --       & {\it source}: --          & {\it source}: trained        & {\it source}: trained \\
              & --       & {\it target}: trained     & {\it target}: {\bf fixed}    & {\it target}: {\bf adapted} \\[1mm]
\textsc{SOM} & --       & {\it source}: --          & {\it source}: trained        & {\it source}: trained \\
              & --       & {\it target}: trained     & {\it target}: {\bf fixed}    & {\it target}: {\bf adapted} \\
\bottomrule
\end{tabular}
\caption{Evaluated systems: \onedim is a one-dimensional (upper) baseline, other systems are multi-di\-men\-sio\-nal.}
\label{tab:systems}
\end{table*}

\begin{table}[htb]
\small
\centering
\begin{tabular}{lccc}
\toprule
{\bf system}  & {\bf SuccRate} & {\bf AvgLen} & {\bf AvgRew} \\
\midrule
\onedim{}     & 97.8\%   & 14.69  & 66.36  \\
\multidim{}   & 97.6\%   & 15.68  & 64.97  \\
\transfixed{} & 96.8\%   & 15.08  & 65.23  \\
\transadapt{} & 97.4\%   & 16.41  & 64.20  \\
\bottomrule
\end{tabular}
\caption{
Test results on simulated data (same error rates as in training): task success rate (SuccRate), average dialogue length (AvgLen), average reward (AvgRew).}
\label{tab:eval_sim-3best}
\end{table}

\subsection{DM Evaluation in Simulation}

To get a better picture of what we might expect during the human evaluation, we first ran evaluations with simulated data. 
The results obtained with the same settings as those during training are shown in \cref{tab:eval_sim-3best}.  As we hypothesised, the scores are very similar, the \onedim system only slightly outperforming the multi-dimensional systems.

\begin{figure*}[htb]
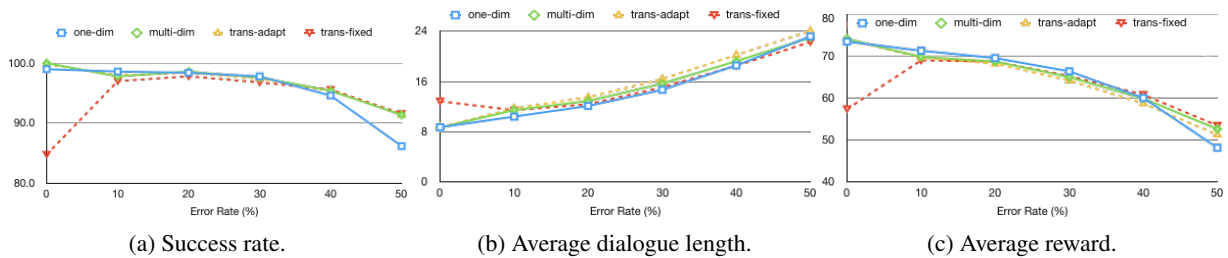

\centering
\begin{subfigure}[b]{.329\textwidth}
\includegraphics[width=\textwidth]{policy_eval_succRate}
\caption{Success rate.}
\label{fig:pol_eval_succRate}
\end{subfigure}
\begin{subfigure}[b]{.329\textwidth}
\includegraphics[width=\textwidth]{policy_eval_avgLen}
\caption{Average dialogue length.}
\label{fig:pol_eval_avgLen}
\end{subfigure}
\begin{subfigure}[b]{.329\textwidth}
\includegraphics[width=\textwidth]{policy_eval_avgRew}
\caption{Average reward.}
\label{fig:pol_eval_avgRew}
\end{subfigure}
\caption{Results in simulation at different error rates.}
\label{fig:pol_eval_sim}
\end{figure*}

We then extended the setup with different semantic error rates \cite{Thomson_ea-SLT-2012}; the results are shown in \cref{fig:pol_eval_sim}.  The performance levels of the four systems are very similar at error rates between 10\% and 40\%, showing that the construction of the multi-dimensional versions in relation to the \onedim baseline is sound, and showing there is no negative transfer, i.e., the adapted systems are not performing worse.\footnote{The discrepancy at zero error rate for the trans-fixed system might have occurred because certain state feature combinations occurring specifically at zero error rate were not seen during training, and might be too distinct to be dealt with by the generalisation capability of the value approximation model used in our reinforcement learning algorithm.}

\section{Evaluation Setup}\label{sec:eval-sys}

We use crowdsourcing to evaluate our system, following \citet{Jurcicek:2011ui} and \citet{Crook:2014kma}.  In both of these works a phone-based system was deployed, using a bespoke ASR and Voice over IP (VoIP) to connect speech input/output with the dialogue system.  Here, we follow a similar evaluation methodology, but with a novel, simpler web-based interface using Google Chrome's built-in web speech API, embedded into the crowdsourcing task webpages. A detailed description of the technical setup can be found in Appendix \ref{sec:sds-setup}.

\subsection{Crowdsourcing Setup}\label{sec:crowds}

The users are recruited on the FigureEight crowdsourcing platform and asked to have a conversation with the system to find a venue meeting certain criteria (e.g.\ cheap Chinese food) and get certain information about that venue (e.g.\ phone number and address).  This scenario is specified in natural language, generated automatically from a set of task specifications randomly generated from the domain ontology.
After each conversation, the user is given a questionnaire to rate the system.

\subsection{Evaluation Metrics}\label{sec:metrics}

The subjective evaluation metrics are derived from the following questionnaire, with one yes/no question (Q1) and four 6-point Likert Scale ratings.
\begin{description}
\small
\item[{Q1 [SubjSucc]:}] Did you find all the information you were looking for? \\[-6mm]
\end{description}
{\small Please state your attitude towards the following statements:}
\begin{description}
\small
\item[{Q2 [VoiceInt]:}] The system was easy to understand (the voice was intelligible). \\[-6mm]
\item[{Q3 [Understand]:}] In this conversation, the system understood what you said. \\[-6mm]
\item[{Q4  [AsExpect]:}] The system worked the way you expected it to during the conversation. \\[-6mm]
\item[{Q5 [WdUseAgain]:}] From your experience with the system, you think you would use it in the future to find a place to eat.
\end{description}

The following objective success metrics are derived from the logs:
\begin{description}
\small
\item \textbf{EntProv}: the system recommended an entity matching the task constraints, \\[-6mm]
\item \textbf{ConstrConf}: the system confirmed all task constraints in its recommendation, \\[-6mm]
\item \textbf{InfoProv}: the system provided all information requested by the user.
\end{description}

\section{Human User Evaluation}\label{sec:results}

\begin{table}[t]
\small
\centering
\begin{tabular}{lcc}
\toprule
{\bf DM version} & {\bf NumDials} & {\bf NumTurns (StDev)} \\
\midrule
\onedim & 245 & 6.67 (2.55) \\
\multidim & 228 & 6.30 (1.97) \\
\transfixed & 261 & 6.57 (2.33) \\
\transadapt & 248 & 6.64 (2.33) \\
\midrule
Total & 982 & 6.55 (2.31) \\
\bottomrule
\end{tabular}
\caption{Corpus statistics: the number of dialogues collected (NumDials) and the average number of turns per dialogue (NumTurns) with standard deviation (StDv).}
\label{tab:stats}
\end{table}

\begin{table*}[htb]
\small
\centering
\begin{tabular}{lcccccccc}
\toprule
\multirow{2}{*}{\bf DM} & {\bf SubjSucc} & {\bf VoiceInt} & {\bf Underst} & {\bf AsExpect} & {\bf WdUseAgain} & \multirow{2}{*}{\bf EntProv} & \multirow{2}{*}{\bf ConstrConf} & \multirow{2}{*}{\bf InfoProv} \\
& {\bf [Q1]} & {\bf [Q2]} & {\bf [Q3]} & {\bf [Q4]} & {\bf [Q5]} & & & \\
\midrule
\onedim     & {\bf 87.3\%} & {\bf 5.49} & {\bf 4.80} & {\bf 4.81} & {\bf 4.67} & {\bf 72.2\%} & {\bf 57.7\%} &      45.7\% \\
\multidim   &      83.3\%       & 5.37  &      4.68  &      4.68       & 4.59  &      68.4\%  &      52.7\%  &      44.7\% \\
\transfixed &      81.6\%       & 5.47  &      4.66  &      4.64       & 4.63  &      70.1\%  &      53.1\%  &      41.0\%  \\
\transadapt &      85.9\%       & 5.38  &      4.67  &      4.64       & 4.57  & {\bf 72.2\%} &      53.1\%  & {\bf 46.6\%} \\
\bottomrule
\end{tabular}
\caption{Overview of subjective and objective evaluation results (cf.~Section~\ref{sec:metrics} for metrics).}
\label{tab:hum-eval}
\end{table*}

\begin{table}[tb]
\small
\centering
\begin{tabular}{lcc}
\toprule
{\bf DM version} & \bf NumDials & {\bf WER} \\
\midrule
\onedim      & 120 & 17.2\% \\
\multidim    & 124 & 15.6\% \\
\transfixed  & 137 & 15.4\% \\
\transadapt  & 115 & {\bf 19.1\%} \\
\bottomrule
\end{tabular}
\caption{WER analysis results (NumDials indicates the number of dialogues transcribed for each system).}
\label{tab:wer-ser}
\end{table}

In total, 982 dialogues were collected (see \cref{tab:stats}), i.e.\ 246 dialogues per system variant on average.  We carried out a number of statistical tests to analyse the observed effect sizes in comparing the systems, including chi-squared (for success rates) and Mann-Whitney tests (for the Likert scale ratings), but also the `two one-sided test', or TOST \cite{Schuirmann-1987}, for \emph{equivalence}, as argued in \cref{sec:models}.
In a TOST scenario, the null hypothesis is that the difference in performance between two systems, $\Delta$, is greater than a given threshold $\epsilon$ (a hyperparameter).  This translates into two one-sided null hypotheses: 
\begin{eqnarray}
H_{\mathrm{lo}}: \Delta \leq -\epsilon \\
H_{\mathrm{hi}}: \Delta \geq +\epsilon
\end{eqnarray}
If both $H_{\mathrm{lo}}$ and $H_{\mathrm{hi}}$ are rejected, we can conclude that $-\epsilon < \Delta < +\epsilon$, i.e. the difference lies below the threshold.  This test is much more conservative than failing to reject the null hypothesis in a conventional statistical test of significant difference. 
The underlying one-sided tests can differ according to the nature of data at hand. The default proposed by \citet{Schuirmann-1987} is t-tests. However, our data fails the normal distribution assumption of a t-test. Therefore, we use the robust t-test of \citet{yuen_approximate_1973} for testing equivalence on Likert scale data, which does not assume normality, and a pooled z-test with continuity correction \cite[p.~53ff.]{fleiss_statistical_2003} for success rates.\footnote{The z statistic is the square root of the $\chi^2$ statistic, which is more suited for determining standard deviation (i.e. size of difference) as opposed to variance.} 
We used a threshold of $\epsilon=10\%$ for the equivalence tests.

\subsection{Evaluation Results}

\Cref{tab:hum-eval} shows the results for both objective and subjective metrics.  When considering the metrics for task success ({\em SubjSucc, EntProv, ConstrConf, InfoProv}), the \onedim system is the highest scoring, although the \transadapt system is often a close second and in some cases the top scorer.  However, no statistically significant differences were detected, and the \onedim system was moreover found to be equivalent to the \multidim ($p=0.024$) and \transadapt ($p=0.002$) systems in perceived success ({\em SubjSucc}), and all three multi-dimensional systems were found to be equivalent to each other ($p=0.006$, $0.009$, and $0.031$).  Similarly, several equivalences were detected for the three objective  success metrics, as illustrated in \cref{sec:equiv}.%
\footnote{Following \citet{armstrong_when_2014}, we do not apply a correction for multiple comparisons \cite{lauzon_easy_2009} since we only performed a limited number of pre-planned comparisons and did not require testing against the universal null hypothesis ``nothing is significant''.}
All systems are equivalent on the other subjective ratings Q2--Q5.

To get a sense of the noise levels encountered by the different system variants, we collected crowdsourced transcriptions of 2,931 utterances from 496 dialogues (45.6\% of the total number of turns in the evaluation corpus and 50.5\% of collected dialogues), spread approximately evenly across all system variants.
We then computed word error rate (WER).
\footnote{The reference transcriptions were obtained by majority voting over the three transcriptions collected for each utterance, with manual fixes in case of a tie (20\% of the utterances).}

Results in \cref{tab:wer-ser} show comparable noise levels for all system variants. 
No significant differences were found and equivalence tests confirmed WER to be equivalent for all the systems.
This confirms that none of the systems was disadvantaged and the results in Table~\ref{tab:hum-eval} are indeed comparable.

\section{Conclusion and Future Work}\label{sec:concl}

In this paper, we have shown that a multi-dimensional, data efficient dialogue manager performs equally to a one-dimensional, more data-hungry (upper) baseline.
In doing so, we have developed a web-based platform for spoken dialogue system evaluation, carried out a crowdsourced user evaluation, and introduced statistical testing for equivalence in our analysis of the results.
All code and data used in our experiments are available at:
\begin{center}\small
\url{https://bitbucket.org/skeizer/madrigal}
\end{center}

The results show that none of the systems outperformed the other systems consistently across various metrics, and more importantly, that several statistical equivalences between the systems could be detected.
We believe that these results are encouraging, especially since we suspect that the use of a web-based speech interface (with inherently varying quality of the microphone used) and the crowdsourcing setup (with inherently varying conditions in which workers do their tasks) resulted in a relatively high level of variance in the data, making it harder to draw strong conclusions.

In the next stage of our research, we aim to further demonstrate the cross-domain transfer capability of the dialogue manager, for example by evaluating partially trained policies, and showing that policies that use transfer learning reach higher performance levels in the early stages of training, or that they achieve a given performance threshold with much less data.

\section*{Acknowledgements}

This research was supported by the EPSRC project MaDrIgAL (EP/N017536/1) and Charles University project PRIMUS/19/SCI/10. 

\bibliography{keizer_etal_sigdial2019}
\bibliographystyle{acl_natbib}

\newpage
\clearpage
\appendix


\section{Dialogue System Setup}\label{sec:sds-setup}

\begin{figure}[tb]
\centering
\includegraphics[width=.35\textwidth]{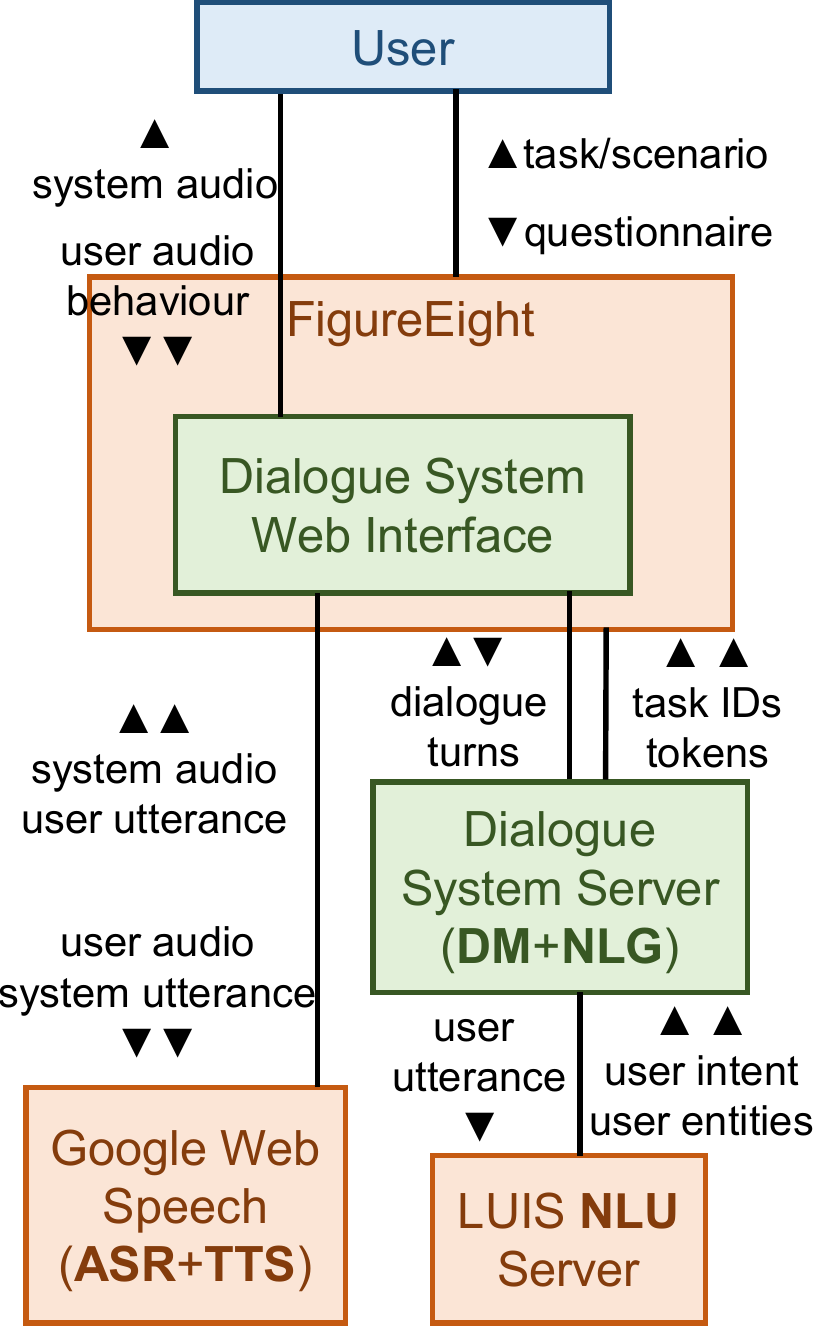}
\caption{Overview of dialogue system evaluation setup.}
\label{fig:eval_sys} 
\end{figure}

An overview of our crowdsourced dialogue system evaluation setup is shown in \cref{fig:eval_sys}.  The core component of the spoken dialogue system is the Dialogue System Server, which contains the DM (see \cref{sec:mdm}), extended with a template-based NLG component and code for processing NLU results from Microsoft's LUIS \cite{Williams_ea-SIGDIAL-2015}.  Our LUIS model was trained with 299 manually constructed and annotated example utterances.

The system is completed by a web-based user interface, which connects with both the Dialogue System Server and the Google Web Speech API.\footnote{\url{https://w3c.github.io/speech-api/speechapi.html}}  User audio input is first sent to Google ASR to get user utterance hypotheses with confidence scores.  These are sent to the Dialogue System Server, which returns a system response utterance.  Finally, this utterance is sent to Google TTS, which returns the synthesised system response audio to be played back to the user.  The web interface is integrated into the FigureEight crowdsourcing platform for managing the evaluation (\cref{sec:crowds}).

\section{Equivalence test results}\label{sec:equiv}

\begin{figure}[tbh]
    \centering
    \begin{subfigure}{.8\linewidth}
    \includegraphics[width=\linewidth]{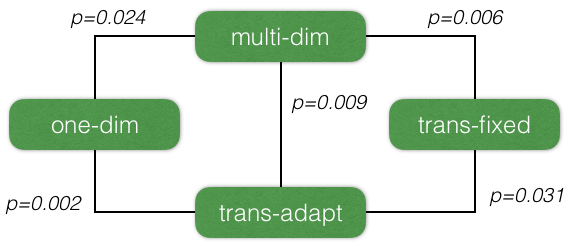}
    \caption{SubjSucc.}
    \label{fig:equiv_SubjSucc}
    \end{subfigure}
    \vspace{2mm}
    \begin{subfigure}{.8\linewidth}
    \includegraphics[width=\linewidth]{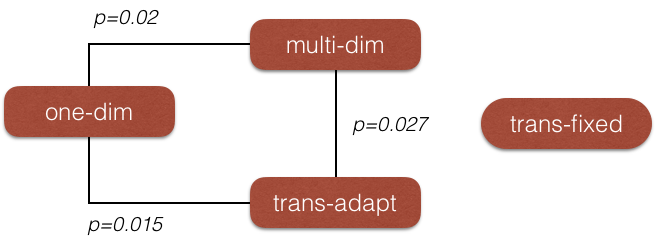}
    \caption{InfoProv.}
    \label{fig:equiv_InfoProv}
    \end{subfigure}
    \vspace{2mm}
    \begin{subfigure}{.8\linewidth}
    \includegraphics[width=\linewidth]{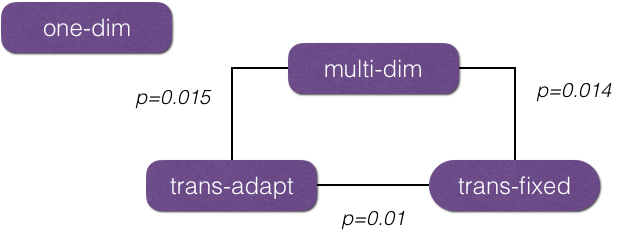}
    \caption{ConstrConf.}
    \label{fig:equiv_ConstrConf}
    \end{subfigure}
    \vspace{2mm}
    \begin{subfigure}{.8\linewidth}
    \includegraphics[width=\linewidth]{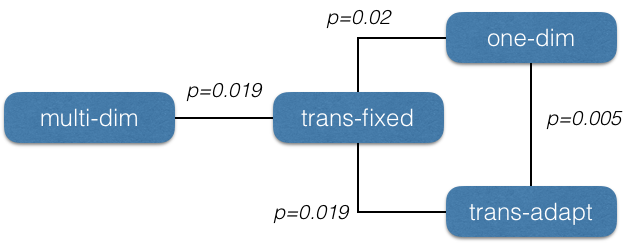}
    \caption{EntProv.}
    \label{fig:equiv_EntProv}
    \end{subfigure}
    \caption{Statistically significant equivalences detected.}
    \label{fig:equiv}
\end{figure}

See Figure~\ref{fig:equiv} for a diagram of all statistically significant equivalences that we detected with respect to the individual evaluation criteria (see Sections~\ref{sec:metrics} and~\ref{sec:results}).

\end{document}